\documentclass[11pt]{article}

\usepackage[preprint]{acl}

\usepackage{times}
\usepackage{latexsym}
\usepackage[T1]{fontenc}
\usepackage[utf8]{inputenc}
\usepackage{microtype}
\usepackage{inconsolata}

\usepackage{graphicx}
\usepackage{amsmath}
\usepackage{amssymb}
\usepackage{url}
\usepackage{booktabs}
\usepackage{array}
\usepackage{tabularx}
\usepackage{multirow}
\usepackage{makecell}
\usepackage{xcolor}
\usepackage{tikz}
\usepackage{cuted}
\usepackage{caption}
\usepackage{pgfplots}
\usetikzlibrary{shapes.geometric}
\usepackage{pifont}
\usepackage{enumitem}
\usepackage{dblfloatfix}
\usepackage{placeins}
\pgfplotsset{compat=1.18}

\IfFileExists{nowidow.sty}{\usepackage[all]{nowidow}}{}
\microtypesetup{protrusion=true, expansion=true}

\renewcommand{\arraystretch}{1.03}
\setlist{nosep,leftmargin=1.25em,topsep=2pt,partopsep=0pt}

\newcolumntype{L}[1]{>{\raggedright\arraybackslash}m{#1}}
\newcolumntype{C}[1]{>{\centering\arraybackslash}m{#1}}
\newcolumntype{Y}{>{\raggedright\arraybackslash}X}
\newcolumntype{Z}{>{\centering\arraybackslash}X}

\graphicspath{{figures/}}
\newcommand{\compacttablefont}{\footnotesize}

\newcommand{\safegraphic}[2][]{%
  \IfFileExists{figures/#2.pdf}{%
    \includegraphics[#1]{figures/#2.pdf}%
  }{%
    \IfFileExists{figures/#2.png}{%
      \includegraphics[#1]{figures/#2.png}%
    }{%
      \fbox{\parbox[c][0.16\textheight][c]{0.92\linewidth}{\centering Missing figure: figures/#2.pdf or figures/#2.png}}%
    }%
  }%
}

\newcommand{\legendentry}[3]{%
\begin{tikzpicture}[baseline=-0.55ex, x=1cm, y=1cm]
  \draw[#1, thick] (0,0) -- (0.36,0);
  #2
  \node[anchor=west, inner sep=1.2pt] at (0.43,0) {#3};
\end{tikzpicture}%
}

\newcommand{\serieslegend}{%
\makebox[\textwidth][c]{%
\begin{minipage}{0.32\textwidth}
\centering
\legendentry{blue}{\fill[blue] (0.18,0) circle (1.2pt);}{RAG-Anything}
\end{minipage}%
\begin{minipage}{0.32\textwidth}
\centering
\legendentry{orange}{\fill[orange] (0.145,-0.035) rectangle (0.215,0.035);}{MMGraphRAG}
\end{minipage}%
\begin{minipage}{0.32\textwidth}
\centering
\legendentry{green!55!black}{\node[regular polygon, regular polygon sides=3, fill=green!55!black, inner sep=1.2pt, rotate=0] at (0.18,0) {};}{TAP-RAG}
\end{minipage}}%
}

\title{TAP-RAG: Task-Aware Policy Control for Long-Document Multimodal Question Answering}
\author{
Zhong Ji\textsuperscript{1},
Keqi Jin\textsuperscript{2},
Yan Zhang\textsuperscript{3,*},
Jiasheng Li\textsuperscript{1,*} \\
\textsuperscript{1}School of Electrical and Information Engineering, Tianjin University \\
\textsuperscript{2}The International Joint Institute of Tianjin University, Fuzhou, Tianjin University \\
\textsuperscript{3}Department of Automation, Tsinghua University \\
\texttt{jizhong@tju.edu.cn},
\texttt{keqi\_jin2002@tju.edu.cn} \\
\texttt{yzhang1995@tsinghua.edu.cn},
\texttt{li\_jiasheng@tju.edu.cn} \\
\textsuperscript{*}Corresponding authors.
}

\begin{document}
\maketitle

\begin{abstract}
Long-document multimodal question answering requires more than retrieving relevant chunks from a large document. Different queries require different evidence behavior. Existing multimodal RAG systems improve evidence access through text chunks, page images, graph links, or heterogeneous document elements, but they often apply a largely query-agnostic evidence-use strategy. We present TAP-RAG, a task-aware policy-controlled RAG framework for long-document multimodal QA. TAP-RAG contains a main controller, the Task-Aware Policy Controller (TAPC), and two policy-guided evidence executors: Task-Aware Query-Guided Flow Diffusion (TA-QFD) and Task-Aware Visual Enhancement (TAVE). For each query, TAPC predicts the task prior, estimates visual/local/global evidence signals, and produces an executable policy. TA-QFD then expands textual and structural evidence over the multimodal document graph, while TAVE selectively inspects page images when visual or layout evidence is needed. A guarded synthesis stage fuses text, visual, and structural evidence and abstains when support is insufficient. On DocBench and MMLongBench-Doc, TAP-RAG achieves the best overall accuracy among the compared systems, improving over a matched multimodal-RAG baseline by $+9.1$ points ($61.1{\to}70.2$) and $+4.5$ points ($42.2{\to}46.7$), respectively. Code is available at \url{https://anonymous.4open.science/r/TAP-RAG}.
\end{abstract}

\section{Introduction}
\label{sec:intro}

Long-document multimodal question answering (QA) requires evidence selection across paragraphs, tables, figures, captions, page layout, metadata, and document structure. A query may ask for a page-local fact, a table cell, a cross-page aggregation, a chart-level relation, a document identifier, or an unsupported claim that should be refused \citep{ma2024mmlongbenchdoc,zou2025docbench}. Retrieval-augmented generation (RAG) improves grounding by providing external evidence to the generator \citep{lewis2020retrieval}. Recent graph-based and multimodal RAG systems further expand retrievable evidence through entities, document units, page images, layout regions, and structural links \citep{edge2024graphrag,guo2025lightrag,wan2025mmgraphrag,guo2025raganything}. However, better access to evidence does not automatically determine how that evidence should be used for a specific query.

\begin{figure}[t]
\centering
\safegraphic[width=\columnwidth]{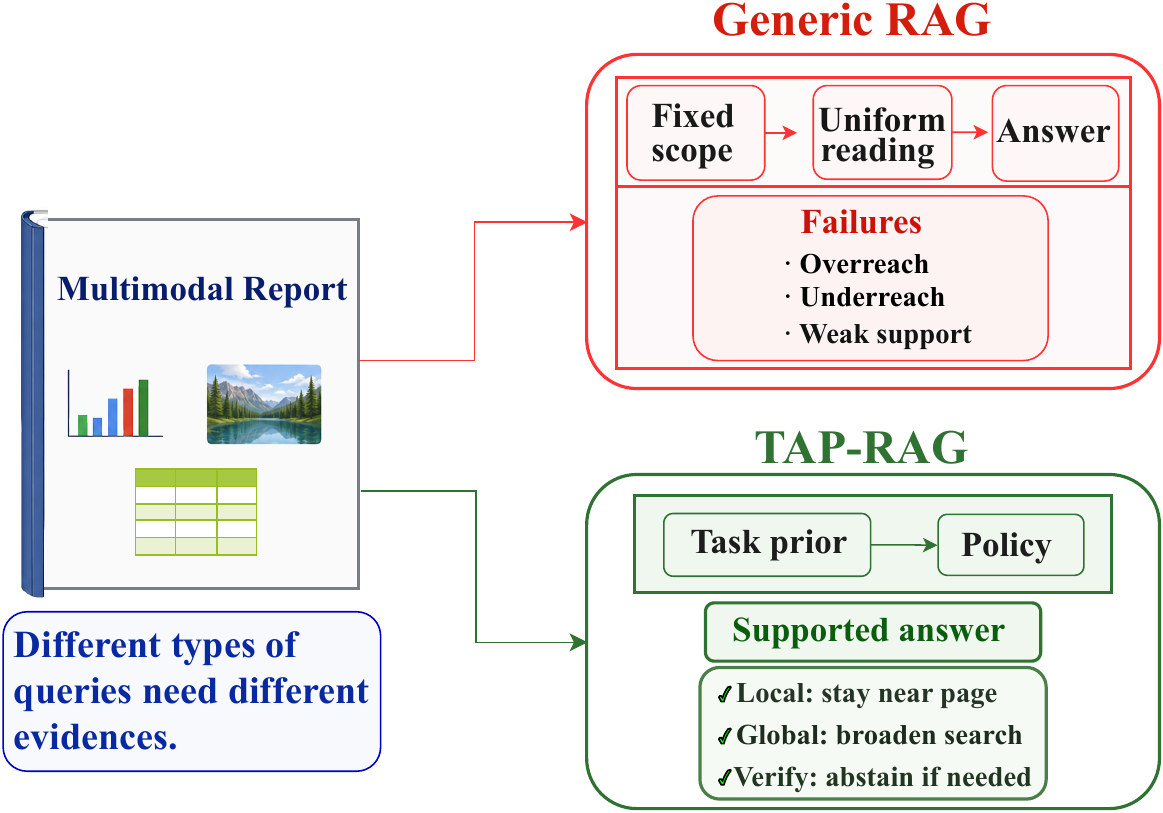}
\caption{Motivating example for policy-controlled multimodal RAG. Generic RAG uses fixed scope and uniform reading, which can cause overreach, underreach, or weak support. TAP-RAG instead uses task priors and policies to adapt evidence use for supported answers.}
\label{fig:motivating_example}
\end{figure}

Figure~\ref{fig:motivating_example} illustrates the motivation. Page-local questions should remain close to page-specific evidence and layout. Table-value questions require row-column alignment rather than only semantic similarity. Aggregation questions need broader traversal across sections or repeated document structures. Visual questions may require selected page-image inspection, but visual evidence should not freely overwrite a well-supported textual answer. Verification and unanswerable questions require stricter support thresholds and conservative synthesis. A query-agnostic pipeline can therefore over-expand local questions, under-explore global ones, over-trust visual guesses, or answer without sufficient support.

We present TAP-RAG, a Task-Aware Policy RAG framework for long-document multimodal QA. TAP-RAG is the overall framework. Inside TAP-RAG, the Task-Aware Policy Controller (TAPC) is the main control module. TAPC does not directly answer the question; instead, it interprets the query, predicts a task prior, estimates visual/local/global evidence needs, and converts them into an executable policy. This policy specifies retrieval locality, seed budget, graph diffusion strength, metadata anchoring, visual acquisition thresholds, fusion authority, structural operators, support thresholds, and abstention behavior.

TA-QFD and TAVE are two sibling evidence-execution modules under TAP-RAG. They are not independent pipelines and are not subparts of TAPC. Rather, both execute the policy produced by TAPC. Task-Aware Query-Guided Flow Diffusion (TA-QFD) operates on the multimodal document graph and expands textual and structural candidates according to the policy's locality, scope, and seed controls. Task-Aware Visual Enhancement (TAVE) selectively inspects page images when the policy indicates visual dependence or when textual evidence is insufficient. In this way, TAPC decides what evidence behavior is needed, while TA-QFD and TAVE acquire the corresponding evidence through graph and visual channels.

Finally, TAP-RAG applies guarded synthesis to combine the outputs of these modules. Evidence from text, visual pages, and structural operators is fused under the policy's fusion and support constraints. If no candidate answer is sufficiently supported, the system abstains instead of forcing an answer. This separation between policy control, graph execution, visual execution, and final synthesis makes TAP-RAG more explicit and auditable than a fixed retrieval-generation pipeline.

\noindent\textbf{Contributions.}
\begin{itemize}
\item We propose TAP-RAG, a task-aware policy-controlled RAG framework for long-document multimodal QA, with TAPC as its core control module for mapping query-level evidence requirements into executable policies.
\item We introduce TA-QFD and TAVE as auxiliary policy-guided modules in TAP-RAG, enabling graph-based evidence expansion and selective visual inspection under TAPC's control.
\item We evaluate TAP-RAG on DocBench and MMLongBench-Doc, showing consistent gains over strong multimodal-RAG baselines and analyzing module interaction through ablations and case studies.
\end{itemize}

\section{Related Work}

\noindent\textbf{Long-document and multimodal document QA.}
Recent benchmarks emphasize cross-page reasoning, visual grounding, fine-grained document understanding, and insufficient-evidence cases \citep{ma2024mmlongbenchdoc,zou2025docbench}. These benchmarks expose mixed evidence requirements: some questions can be answered from one local span, while others require table structure, image layout, or document-level comparison. TAP-RAG targets this mixed setting by using one multimodal document representation while varying query-time evidence behavior through an explicit policy.

\begin{figure*}[!t]
\centering
\safegraphic[width=0.98\textwidth]{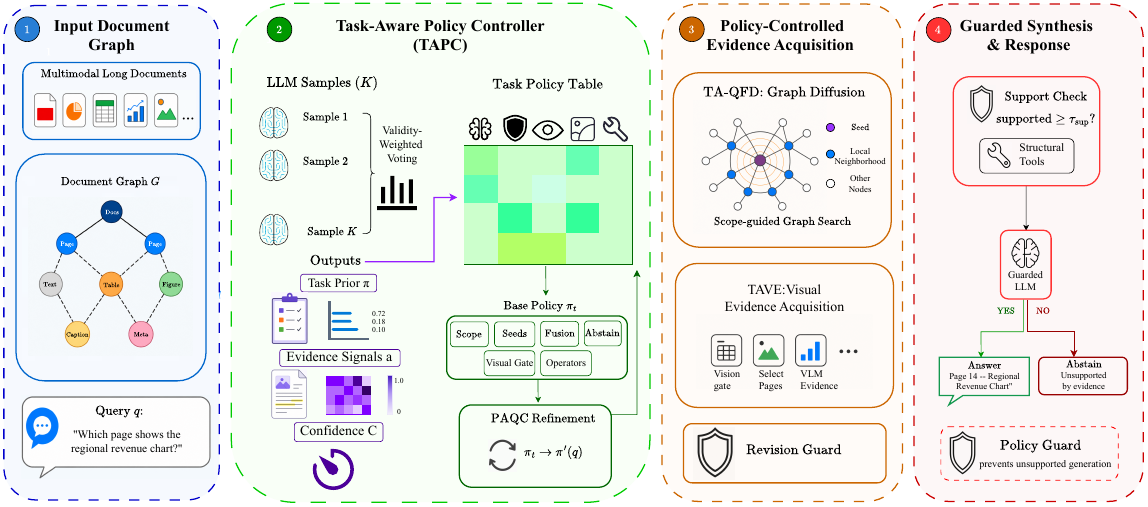}
\caption{Overview of TAP-RAG. TAPC produces the task prior, evidence signals, confidence, and executable policy. TA-QFD and TAVE acquire graph and visual evidence under the policy, and guarded synthesis prevents unsupported generation.}
\label{fig:taprag_framework}
\end{figure*}

\noindent\textbf{Multimodal representation and document RAG.}
RAG grounds language-model outputs in external evidence \citep{lewis2020retrieval}. GraphRAG uses graph communities for query-focused retrieval \citep{edge2024graphrag}; LightRAG combines graph and vector retrieval \citep{guo2025lightrag}; MMGraphRAG extends graph retrieval to multimodal document QA \citep{wan2025mmgraphrag}; and systems such as ColPali, VisRAG, M3DocRAG, and RAG-Anything add page-image retrieval or heterogeneous document-element handling \citep{faysse2024colpali,yu2024visrag,cho2024m3docrag,guo2025raganything}. Orthogonal MLLM-design work improves modality coordination and image-text alignment through modality-expert adaptation, momentum-contrast semantic enhancement, and keyword-explicit reasoning \citep{mecaModalityExperts,userMomentumContrast,eliminateBeforeAlign}. TAP-RAG is complementary: it focuses on how retrieved evidence should be expanded, inspected, fused, or rejected for each query.

\noindent\textbf{Routing and abstention.}
Self-consistency improves reasoning by aggregating multiple samples \citep{wang2023selfconsistency}. Adaptive RAG methods decide whether retrieval is needed or route among coarse retrieval-generation paths \citep{asai2024selfrag,yan2024crag,jeong2024adaptiverag}. TAP-RAG uses sampling inside TAPC, but maps the routed task prior and evidence signals into field-level controls over diffusion, visual acquisition, fusion, structural operators, support checking, and abstention.

\section{Method}
\label{sec:method}

Figure~\ref{fig:taprag_framework} summarizes TAP-RAG. The framework assumes a parsed multimodal document graph and focuses on query-time control. TAPC first resolves the evidence behavior required by the query, then converts it into an executable policy. The policy controls graph expansion, visual acquisition, fusion authority, structural operators, and support checking. This design makes the interaction between modules explicit: TAPC decides the behavior, TA-QFD and TAVE acquire evidence under that behavior, and guarded synthesis determines whether the final answer is sufficiently supported.

\subsection{Document Graph and Policy}

\noindent\textbf{Multimodal document graph.}
Let $q$ be a query and $G=(V,E)$ a multimodal document graph:
\begin{equation}
\label{eq:node_sets}
V = V_p \cup V_t \cup V_{\mathrm{tab}} \cup V_f \cup V_m,
\end{equation}
where $V_p$, $V_t$, $V_{\mathrm{tab}}$, $V_f$, and $V_m$ denote page, text-unit, table-region, figure/image, and metadata-anchor nodes. The edge set is
\begin{equation}
\label{eq:edge_sets}
\begin{aligned}
E ={}& E_{\mathrm{contain}} \cup E_{\mathrm{next}} \cup E_{\mathrm{caption}}
      \cup E_{\mathrm{semantic}} \\
&\cup E_{\mathrm{layout}} \cup E_{\mathrm{cross}},
\end{aligned}
\end{equation}
covering containment, sequence, caption, semantic, layout, and cross-modal relations. The graph therefore represents both textual continuity and document structure, allowing the controller to choose between local and broad evidence traversal.

\begin{table*}[t]
\centering
\begingroup
\compacttablefont
\setlength{\tabcolsep}{3.0pt}
\renewcommand{\arraystretch}{1.08}
\begin{tabularx}{\textwidth}{@{}L{2.0cm}C{0.95cm}C{1.0cm}L{2.7cm}Y@{}}
\toprule
Field & Type & Mut. & Controlled stage & Runtime role \\
\midrule
$\alpha_{\mathrm{diff}}$ & cont. & yes & Graph diffusion & Sets restart/locality strength: larger values keep evidence near seeds, while smaller values allow broader traversal. \\
$K_{\mathrm{seed}}$ & int. & yes & Seed selection & Controls how many semantic and structural seed nodes initialize diffusion. \\
$B_{\mathrm{scope}}$ & cont. & yes & Scope protection & Penalizes cross-document or off-scope evidence. \\
$B_{\mathrm{meta}}$ & cont. & task & Metadata anchoring & Boosts title-page, header, author, and document-identifying anchors. \\
$\tau_{\mathrm{text}}$ & cont. & yes & Text gate & Determines whether text evidence is sufficient without visual inspection. \\
$\tau_{\mathrm{acq}}$ & cont. & yes & Visual acquisition & Triggers page-image inspection when visual evidence is likely needed. \\
$\tau_{\mathrm{sup}}$ & cont. & yes & Support checking & Sets the minimum support required before accepting an answer. \\
$f_{\mathrm{fusion}}$ & disc. & constr. & Cross-modal fusion & Chooses whether visual evidence validates, supplements, or revises text evidence. \\
$a_{\mathrm{abs}}$ & cont. & yes & Abstention & Raises the refusal prior when evidence support is weak. \\
$O_{\mathrm{struct}}$ & set & constr. & Structural execution & Enables page lookup, metadata extraction, scoped counting, or table-cell validation. \\
\bottomrule
\end{tabularx}
\endgroup
\caption{Policy vector fields used by TAP-RAG. Each component links task interpretation to concrete execution behavior.}
\label{tab:policy_fields}
\end{table*}

\noindent\textbf{Policy definition.}
A policy is a structured execution vector
\begin{equation}
\label{eq:policy_vector}
\begin{split}
\pi(q)=(&\alpha_{\mathrm{diff}},K_{\mathrm{seed}},B_{\mathrm{scope}},B_{\mathrm{meta}},\\
&\tau_{\mathrm{text}},\tau_{\mathrm{acq}},\tau_{\mathrm{sup}},f_{\mathrm{fusion}},\\
&a_{\mathrm{abs}},O_{\mathrm{struct}}).
\end{split}
\end{equation}
Table~\ref{tab:policy_fields} lists the policy fields. Continuous fields tune retrieval locality and thresholds; discrete fields constrain fusion and structural execution. Given $q$, TAPC resolves a task prior $t$, evidence signals $a=(a^v,a^l,a^g)$, and confidence $C$. The task prior indexes the Task Policy Table, $\pi_t=\mathrm{TPT}(t)$, and bounded query control refines it into $\pi'(q)$.

\subsection{Task-Aware Policy Controller (TAPC)}

TAPC is the control layer of TAP-RAG. It routes the query to a TPT category, aggregates evidence-need signals, and converts the result into executable controls for TA-QFD, TAVE, and guarded synthesis. This separates task interpretation from evidence execution: the model does not merely retrieve more context, but decides what kind of evidence behavior is needed before retrieval expansion and visual inspection.

\noindent\textbf{Task Policy Table (TPT).}
The TPT contains seven categories: metadata lookup, evidence localization, document aggregation, value extraction, visual reasoning, Boolean verification, and generic reasoning. Each category defines a base evidence-use behavior. For example, metadata lookup emphasizes title-page and header anchors, value extraction favors local table or numeric evidence, and aggregation allows broader traversal. Ambiguous queries are further regulated by evidence signals and routing confidence instead of relying only on a hard label.

\noindent\textbf{Validity-weighted voting.}
TAPC samples $K$ LLM completions $s_i=(t_i,z_i,r_i,h_i,a_i)$, where $t_i$ is the TPT category, $z_i$ stores interpretation fields, $r_i$ is a rationale, $h_i$ contains optional hints, and $a_i\in[0,1]^3$ estimates visual, local, and global evidence needs. Each sample receives
\begin{equation}
\label{eq:validity}
\begin{aligned}
v_i ={}& w_sS_{\mathrm{sem}}(s_i)
+w_fS_{\mathrm{field}}(s_i)\\
&+w_gS_{\mathrm{ground}}(s_i),
\quad w_s+w_f+w_g=1 .
\end{aligned}
\end{equation}
$S_{\mathrm{sem}}$ checks task-answer compatibility, $S_{\mathrm{field}}$ measures agreement with neighboring samples, and $S_{\mathrm{ground}}$ checks whether the predicted exact term is grounded in the query.

The final task prior and evidence signals are
\begin{equation}
\label{eq:voting}
\begin{aligned}
t &=
\arg\max_{\ell}\sum_i v_i\,\mathbb{I}[t_i=\ell],\\
a^m &=
\frac{\sum_i v_i a_i^m}{\sum_i v_i},
\qquad m\in\{v,l,g\}.
\end{aligned}
\end{equation}
Confidence combines vote concentration and signal consistency:
\begin{equation}
\label{eq:confidence}
\begin{aligned}
C &=
\lambda_t C_t + \lambda_a C_a,
\qquad \lambda_t+\lambda_a=1,\\
C_t &=
\frac{\max_{\ell}\sum_i v_i\,\mathbb{I}[t_i=\ell]}{\sum_i v_i},\\
C_a &=
1-\frac{1}{3}\sum_{m\in\{v,l,g\}}
\operatorname{NormVar}(\{a_i^m\}_{i=1}^{K}).
\end{aligned}
\end{equation}
$\operatorname{NormVar}$ is clipped to $[0,1]$. High confidence allows stronger query-level refinement, while low confidence keeps the policy closer to the conservative TPT default.

\begin{table*}[t]
\centering
\begingroup
\compacttablefont
\begin{tabularx}{\textwidth}{@{}L{3.0cm}L{2.6cm}L{2.4cm}L{2.7cm}Y@{}}
\toprule
Task category & Retrieval scope & Visual prior & Fusion action & Structural operators \\
\midrule
Metadata lookup & anchor-local & conditional & text-validate & metadata and page operators \\
Evidence localization & local/scope & conditional & conservative & page and section operators \\
Document aggregation & broad & low/conditional & text-first & scoped count when complete \\
Value extraction & local & conditional/high & cross-validate & parser operators if structured \\
Visual reasoning & visual-local & high & visual-gated & disabled by default \\
Boolean verification & claim-scope & conditional & conservative & optional if structured \\
Generic reasoning & adaptive & conditional & balanced & disabled by default \\
\bottomrule
\end{tabularx}
\endgroup
\caption{Compact Task Policy Table (TPT) inside TAPC. Each row records base evidence-use behavior, which bounded query control refines using visual, local, and global evidence signals.}
\label{tab:task_policy_table}
\end{table*}

\noindent\textbf{Bounded query control.}
Table~\ref{tab:task_policy_table} defines the base policy row for each TPT category. TAPC computes $\pi'(q)=\pi_t\oplus\Delta(a,C)$, where $\oplus$ denotes clip-then-substitute composition. For a mutable continuous field $j$,
\begin{equation}
\label{eq:paqc_cont}
\begin{aligned}
\pi'_j ={}& \pi_{t,j}
+M_{t,j} C\,\operatorname{clip}(\eta_j g_j(a,t),\\
&-\epsilon_j,\epsilon_j).
\end{aligned}
\end{equation}
$M_{t,j}$ is the mutability mask, $\eta_j$ is a step size, $\epsilon_j$ bounds perturbation, and $g_j(a,t)$ gives the control direction. Discrete fields choose the highest-scoring allowed action:
\begin{equation}
\label{eq:action_score}
\begin{aligned}
\operatorname{Score}(r) ={}& \omega_1\operatorname{Compat}(r,\pi_t)
+\omega_2\operatorname{EviFit}(r,a)\\
&-\omega_3\operatorname{Risk}(r,t)(1-C).
\end{aligned}
\end{equation}
The selected fusion action is $f'_{\mathrm{fusion}}=\arg\max_{r\in\mathcal{A}_t}\operatorname{Score}(r)$. Locked fields remain unchanged, so query-level adaptation cannot override hard scope or safety constraints.

\subsection{Policy-Controlled Evidence Acquisition}

\noindent\textbf{Task-Aware Query-Guided Flow Diffusion (TA-QFD).}
Task-Aware Query-Guided Flow Diffusion conditions seed budget, locality, node priors, edge weights, and scope protection on $\pi'(q)$. From semantic-structural seeds $s$, each node $v$ receives
\begin{equation}
\label{eq:node_prior}
\begin{aligned}
r_v ={}& \mu_1\operatorname{sim}(q,v)
+\mu_2\operatorname{lex}(q,v)\\
&+\mu_3 b_v(\pi'(q)).
\end{aligned}
\end{equation}
Here $b_v$ boosts nodes such as page-hint neighborhoods, metadata anchors, visual regions, or global evidence depending on the policy. This allows the same graph to support local lookup, table extraction, and document aggregation without changing the underlying index.

Edges are reweighted as $w'_{uv}=w_{uv}\cdot A_{uv}\cdot B_{uv}$:
\begin{equation}
\label{eq:edge_reweight}
\begin{aligned}
A_{uv} &= \exp\!\left(
\gamma\left(\frac{\operatorname{rel}(u)+\operatorname{rel}(v)}{2}-0.3\right)
\right),\\
B_{uv} &= \max\bigl(
0.01,1-\sigma\,\mathbb{I}[\operatorname{crossdoc}(u,v)]
\bigr).
\end{aligned}
\end{equation}
With row-normalized transition $P'$, diffusion is
\begin{equation}
\label{eq:diffusion}
\begin{aligned}
p^{(k+1)}={}&\alpha_{\mathrm{diff}}s\\
&+(1-\alpha_{\mathrm{diff}}){P'}^\top p^{(k)} .
\end{aligned}
\end{equation}
Larger $\alpha_{\mathrm{diff}}$ keeps evidence local; smaller values allow broader traversal. Thus, TA-QFD improves candidate coverage while preserving the controller's decision about whether the query should remain local or expand globally.

\noindent\textbf{Task-Aware Visual Enhancement (TAVE).}
Task-Aware Visual Enhancement first produces a text-path candidate $a_t$ and estimates textual sufficiency:
\begin{equation}
\label{eq:text_suff}
\begin{aligned}
S_{\mathrm{text}} ={}&
a_1\operatorname{Coverage}(q,E_t)
+a_2\operatorname{Rel}(q,E_t)\\
&+a_3\operatorname{Support}(a_t,E_t).
\end{aligned}
\end{equation}
The VLM is invoked when $S_{\mathrm{text}}<\tau_{\mathrm{text}}$ or $a^v>\tau_{\mathrm{acq}}$. Candidate pages are scored by
\begin{equation}
\label{eq:page_score}
\begin{aligned}
P(p) ={}& b_1\operatorname{RetrievalPage}(p)
+b_2\operatorname{HintMatch}(p,h)\\
&+b_3\operatorname{MetaAnchor}(p)
+b_4a^v\operatorname{VisPrior}(p).
\end{aligned}
\end{equation}
For visual candidate $a_v$, revision is allowed only when visual grounding is strong and text support is weak:
\begin{equation}
\label{eq:revision}
\begin{aligned}
\operatorname{Revise}(a_t,a_v)=\mathbb{I}[&
R_v>\tau_{\mathrm{rel}}\land S_v>\tau_{\mathrm{sup}}\\
&\land Q(a_v)>Q(a_t)+\epsilon_q\\
&\land f'_{\mathrm{fusion}}\in\mathcal{A}_{\mathrm{revise}}].
\end{aligned}
\end{equation}
Otherwise, visual evidence validates or supplements the text path without freely overwriting it. This guarded behavior is important for visually dense PDFs, where plausible visual interpretations can conflict with table structure or extracted text.

\subsection{Guarded Synthesis}

For each candidate $a$, TAP-RAG scores
\begin{equation}
\label{eq:candidate_quality}
\begin{aligned}
Q(a)={}&d_1S_{\mathrm{sup}}(a,E)
+d_2S_{\mathrm{comp}}(a,q)\\
&+d_3S_{\mathrm{spec}}(a).
\end{aligned}
\end{equation}
Support is channel-aware:
\begin{equation}
\label{eq:channel_support}
\begin{aligned}
S_{\mathrm{sup}}(a,E)=\max\{&
S^{\mathrm{text}}_{\mathrm{sup}}(a,E_t),
S^{\mathrm{vis}}_{\mathrm{sup}}(a,E_v),\\
&S^{\mathrm{struct}}_{\mathrm{sup}}(a,E_s)\}.
\end{aligned}
\end{equation}
Text support comes from retrieved spans, visual support from inspected pages or regions, and structural support from deterministic operators such as table-cell or page validation. If no candidate meets the policy-adjusted support threshold, TAP-RAG abstains. This final step closes the loop between policy and execution: retrieval and visual inspection propose evidence, but the answer is accepted only when the selected policy's support requirement is satisfied.

\section{Experiments}
\label{sec:experiments}

\subsection{Experimental Setup}

\noindent\textbf{Benchmarks and metrics.}
We evaluate on DocBench and MMLongBench-Doc. DocBench covers academic, financial, government, legal, and news PDFs with text-only, multimodal, and unanswerable questions \citep{zou2025docbench}. MMLongBench-Doc covers long-context document understanding over reports, tutorials, papers, guidebooks, brochures, administration files, and financial reports \citep{ma2024mmlongbenchdoc}. We report answer accuracy in percent and use the same benchmark splits and page-range buckets across systems.

\noindent\textbf{Compared systems.}
The comparison includes GPT-4o-mini \citep{openai2024gpt4omini}, Qwen3-VL-Plus \citep{qwen2025qwen3vl}, GraphRAG \citep{edge2024graphrag}, LightRAG \citep{guo2025lightrag}, MMGraphRAG \citep{wan2025mmgraphrag}, and RAG-Anything \citep{guo2025raganything}. GPT-4o-mini and Qwen3-VL-Plus are direct model baselines. GraphRAG uses GPT-4o while lightrag uses GPT-4o-mini as the generation backbone. MMGraphRAG is a published-reference result from the RAG-Anything paper under its GPT-4o-mini setting. RAG-Anything is evaluated with Qwen3-VL-Plus as the backbone, and TAP-RAG uses the same Qwen3-VL-Plus backbone for generation and page-image vision calls.

\noindent\textbf{Diagnostic view.}
Domain and length results measure overall robustness. Since TAP-RAG's central claim is query-level evidence control, mechanism-level evidence is provided by the TAPC/TA-QFD/TAVE ablation and case studies. This avoids treating domain labels as a proxy for task type and instead evaluates whether the controller and executors interact as intended.

\subsection{Domain-Level Accuracy}
\label{sec:domain_results}

\begin{table*}[!t]
\centering
\begingroup
\compacttablefont
\setlength{\tabcolsep}{1.7pt}
\renewcommand{\arraystretch}{0.92}
\newcommand{\methodcite}[2]{\mbox{#1~\citep{#2}}}
\newcommand{\raganythingmethod}{\makecell[l]{RAG-Anything~\citep{guo2025raganything}\\(Qwen3-VL-Plus backbone)}}
\begin{tabularx}{\textwidth}{@{}L{5.3cm}*{9}{Z}@{}}
\toprule
\multicolumn{10}{c}{DocBench} \\
\midrule
Method & Aca. & Fin. & Gov. & Law & News & Txt. & Mm. & Una. & All \\
\midrule
\methodcite{GPT-4o-mini}{openai2024gpt4omini} & 40.3 & 46.9 & 60.3 & 59.2 & 61.0 & 61.0 & 43.8 & 49.6 & 51.2 \\
\methodcite{Qwen3-VL-Plus}{qwen2025qwen3vl} & 53.4 & 45.0 & 60.0 & 59.4 & 62.1 & 63.9 & 56.8 & 43.1 & 54.5 \\
\methodcite{GraphRAG}{edge2024graphrag} & 40.6 & 27.1 & 56.8 & 59.7 & 75.0 & 73.5 & 24.4 & 76.6 & 54.7 \\
\methodcite{LightRAG}{guo2025lightrag} & 53.8 & 56.2 & 59.5 & 61.8 & 65.7 & 85.0 & 59.7 & 46.8 & 58.4 \\
\methodcite{MMGraphRAG}{wan2025mmgraphrag} & 64.3 & 52.8 & 64.9 & 40.0 & 61.5 & 67.6 & 66.0 & 60.5 & 61.0 \\
\raganythingmethod & 64.1 & 59.2 & 55.8 & 54.0 & 68.9 & 72.1 & \textbf{69.5} & \textbf{87.1} & 61.1 \\
TAP-RAG & \textbf{67.1} & \textbf{64.3} & \textbf{75.0} & \textbf{74.7} & \textbf{76.2} & \textbf{87.4} & 64.9 & 71.8 & \textbf{70.2} \\
\bottomrule
\end{tabularx}

\vspace{0.95\baselineskip}

\begin{tabularx}{\textwidth}{@{}L{5.3cm}*{8}{Z}@{}}
\toprule
\multicolumn{9}{c}{MMLongBench-Doc} \\
\midrule
Method & Res. & Tut. & Acad. & Guid. & Bro. & Adm. & Fin. & All \\
\midrule
\methodcite{GPT-4o-mini}{openai2024gpt4omini} & 35.5 & 44.0 & 24.6 & 33.1 & 29.5 & 46.8 & 31.1 & 33.5 \\
\methodcite{Qwen3-VL-Plus}{qwen2025qwen3vl} & 39.2 & 37.1 & 34.6 & 38.6 & 31.7 & 35.9 & 34.0 & 36.5 \\
\methodcite{GraphRAG}{edge2024graphrag} & 30.8 & 27.0 & 25.0 & 29.7 & 24.0 & 34.4 & 16.7 & 27.2 \\
\methodcite{LightRAG}{guo2025lightrag} & 40.8 & 34.1 & 36.2 & 39.4 & \textbf{41.0} & 44.4 & 38.3 & 38.9 \\
\methodcite{MMGraphRAG}{wan2025mmgraphrag} & 40.8 & 36.5 & 35.7 & 35.8 & 28.2 & 46.9 & 38.5 & 37.7 \\
\raganythingmethod & 46.6 & 43.5 & 38.7 & 43.9 & 34.0 & 45.7 & 43.6 & 42.2 \\
TAP-RAG & \textbf{50.9} & \textbf{47.4} & \textbf{43.9} & \textbf{48.9} & 38.5 & \textbf{49.7} & \textbf{47.6} & \textbf{46.7} \\
\bottomrule
\end{tabularx}
\endgroup
\caption{Domain-level answer accuracy on DocBench and MMLongBench-Doc. Bold values mark the best result in each column. MMGraphRAG uses GPT-4o-mini as its backbone, while RAG-Anything and TAP-RAG use Qwen3-VL-Plus.}
\label{tab:domain_results}
\end{table*}

Table~\ref{tab:domain_results} shows that TAP-RAG achieves the best overall accuracy on both benchmarks. Compared with the strongest same-backbone baseline, RAG-Anything with Qwen3-VL-Plus, TAP-RAG improves by $+9.1$ points on DocBench and $+4.5$ points on MMLongBench-Doc. The gains span many domains, especially government, legal, news, research, tutorial, academic, guidebook, administration, and financial documents.

These results support the motivation of TAP-RAG: long-document multimodal QA requires controlled evidence use, not only more retrieval. TAPC selects a task-aware policy before evidence acquisition, adjusting retrieval scope, visual inspection, fusion, and abstention before answer synthesis. The DocBench multimodal and unanswerable columns remain challenging, suggesting room for stronger visual recall and refusal calibration.

\subsection{Performance by Document Length}
\label{sec:page_range}

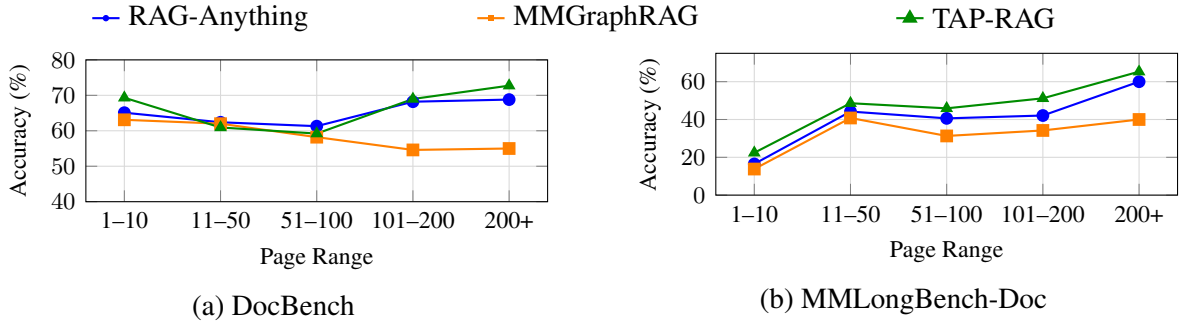
\begin{figure*}[!t]
\centering
\serieslegend
\vspace{0.15\baselineskip}
\begin{minipage}{0.48\textwidth}
\centering
\begin{tikzpicture}
\begin{axis}[
  width=\linewidth,
  height=0.45\linewidth,
  ymin=40,ymax=80,
  ytick={40,50,60,70,80},
  ylabel={Accuracy (\%)},
  xlabel={Page Range},
  symbolic x coords={1--10,11--50,51--100,101--200,200+},
  xtick=data,
  tick label style={font=\small},
  label style={font=\small},
  grid=major,
  grid style={gray!30},
]
\addplot[blue, mark=*, thick] coordinates {(1--10,65.1) (11--50,62.4) (51--100,61.3) (101--200,68.2) (200+,68.8)};
\addplot[orange, mark=square*, thick] coordinates {(1--10,63.1) (11--50,62.0) (51--100,58.2) (101--200,54.6) (200+,55.0)};
\addplot[green!55!black, mark=triangle*, thick] coordinates {(1--10,69.32) (11--50,60.93) (51--100,59.21) (101--200,68.97) (200+,72.73)};
\end{axis}
\end{tikzpicture}
(a) DocBench
\end{minipage}\hfill
\begin{minipage}{0.48\textwidth}
\centering
\begin{tikzpicture}
\begin{axis}[
  width=\linewidth,
  height=0.45\linewidth,
  ymin=0,ymax=75,
  ytick={0,20,40,60},
  ylabel={Accuracy (\%)},
  xlabel={Page Range},
  symbolic x coords={1--10,11--50,51--100,101--200,200+},
  xtick=data,
  tick label style={font=\small},
  label style={font=\small},
  grid=major,
  grid style={gray!30},
]
\addplot[blue, mark=*, thick] coordinates {(1--10,16.5) (11--50,44.2) (51--100,40.6) (101--200,42.1) (200+,60.0)};
\addplot[orange, mark=square*, thick] coordinates {(1--10,13.8) (11--50,40.8) (51--100,31.3) (101--200,34.2) (200+,40.0)};
\addplot[green!55!black, mark=triangle*, thick] coordinates {(1--10,22.4) (11--50,48.6) (51--100,45.9) (101--200,51.2) (200+,65.3)};
\end{axis}
\end{tikzpicture}
(b) MMLongBench-Doc
\end{minipage}
\caption{Accuracy by document page range for TAP-RAG, RAG-Anything, and MMGraphRAG under the same page-range buckets.}
\label{fig:page_range}
\end{figure*}

Figure~\ref{fig:page_range} evaluates whether the method remains useful as documents become longer. TAP-RAG is competitive across all page ranges and is strongest in the longest MMLongBench-Doc buckets, where relevant evidence often spans headings, tables, captions, and explanatory text. This pattern is consistent with the design of TAPC and TA-QFD: longer documents require broader but still bounded evidence traversal, while the policy prevents expansion from drifting into irrelevant pages. TAVE further helps when long reports contain visually organized tables or page-level cues that are difficult to recover from extracted text alone. On DocBench, medium-length documents remain difficult because they contain enough distractors to confuse retrieval but fewer repeated structures for diffusion to reinforce. Exact page-range values are listed in Appendix Table~\ref{tab:page_values_app}.

\subsection{Ablation and Executor Coupling}
\label{sec:ablation}

Table~\ref{tab:executor_coupling_ablation} evaluates TAPC and the two policy-guided executors. The goal is to test interaction rather than a purely additive module stack: TA-QFD expands graph candidates, while TAVE verifies selected pages visually. The full system uses both under the same policy and then applies support-gated synthesis. This setting directly tests whether executor behavior remains beneficial when evidence expansion, visual checking, and final acceptance are governed by one shared controller.

\begin{table*}[!t]
\centering
\begingroup
\compacttablefont
\setlength{\tabcolsep}{1.8pt}
\renewcommand{\arraystretch}{0.96}
\begin{tabularx}{\textwidth}{@{}L{2.35cm}*{6}{Z}*{8}{Z}@{}}
\toprule
\multirow{2}{*}{Variant}
& \multicolumn{6}{c}{DocBench}
& \multicolumn{8}{c}{MMLongBench-Doc} \\
\cmidrule(lr){2-7}\cmidrule(l){8-15}
& Aca. & Fin. & Gov. & Law & News & All
& Res. & Tut. & Acad. & Guid. & Bro. & Adm. & Fin. & All \\
\midrule
Baseline
& 64.1 & 59.2 & 55.8 & 54.0 & 68.9 & 61.1
& 46.6 & 43.5 & 38.7 & 43.9 & 34.0 & 45.7 & 43.6 & 42.2 \\
TAPC only
& 66.6 & 63.9 & 74.3 & 73.9 & 75.0 & 69.6
& 50.1 & 47.1 & 43.5 & 48.6 & 37.7 & 49.3 & 47.2 & 46.6 \\
TAPC + TA-QFD
& 66.9 & 63.2 & 74.9 & 74.1 & 74.3 & 69.1
& 50.7 & 46.8 & 43.8 & 47.9 & 36.8 & 49.0 & 46.6 & 46.1 \\
TAPC + TAVE
& 65.8 & 64.1 & 73.3 & 72.8 & 75.8 & 68.7
& 48.7 & 45.9 & 42.8 & 48.8 & 38.2 & 48.4 & 47.4 & 45.5 \\
Full TAP-RAG
& \textbf{67.1} & \textbf{64.3} & \textbf{75.0} & \textbf{74.7} & \textbf{76.2} & \textbf{70.2}
& \textbf{50.9} & \textbf{47.4} & \textbf{43.9} & \textbf{48.9} & \textbf{38.5} & \textbf{49.7} & \textbf{47.6} & \textbf{46.7} \\
\bottomrule
\end{tabularx}
\endgroup
\caption{Domain-level ablation of TAPC and the two policy-guided evidence executors on DocBench and MMLongBench-Doc. Full TAP-RAG couples TA-QFD candidate expansion, TAVE page verification, and guarded synthesis.}
\label{tab:executor_coupling_ablation}
\end{table*}

TAPC only is already strong, showing that query-level policy resolution is the main source of control. It adjusts locality, visual-trigger thresholds, fusion authority, structural operators, and support requirements before generation. However, TAPC alone cannot recover evidence missing from the initial candidate set, and it cannot directly inspect layout-sensitive pages.

TA-QFD improves candidate coverage through graph diffusion, which benefits text-heavy and structure-heavy documents. Yet expansion can also introduce semantically related distractors when the query is highly local. TAVE provides the complementary behavior by verifying candidate pages visually, but it depends on the quality of candidate pages. If retrieval misses the correct page or includes visually plausible distractors, visual inspection alone can still be misled.

The full model is strongest because the modules form a controlled loop. TAPC selects the policy, TA-QFD expands evidence under the selected locality and scope constraints, TAVE verifies layout or visual evidence only when needed, and guarded synthesis prevents noisy graph or visual candidates from dominating the final answer. This interaction explains why the two executors are most effective when coordinated rather than used as independent add-ons.

\begin{figure*}[!t]
\centering
\safegraphic[width=0.98\textwidth,height=0.285\textheight,keepaspectratio]{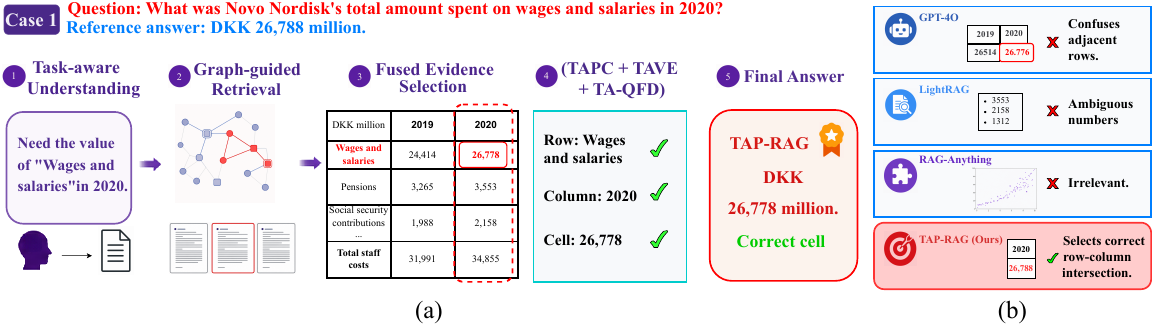}
\caption{Case study on table-cell value extraction from case1.pdf. (a) Baselines are distracted by adjacent rows or ambiguous numbers. (b) TAP-RAG follows a task-aware evidence path, verifies the row-column intersection, and returns DKK 26,778 million.}
\label{fig:case1}

\vspace{0.75\baselineskip}

\safegraphic[width=0.98\textwidth,height=0.285\textheight,keepaspectratio]{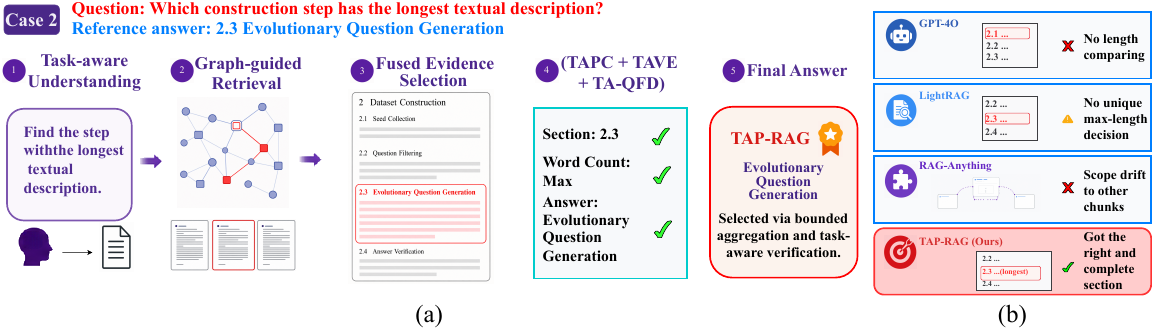}
\caption{Case study on section-level textual aggregation from case2.pdf. (a) Baselines retrieve related sections but do not make a reliable maximum-length decision. (b) TAP-RAG performs bounded aggregation over candidate sections and selects Section 2.3, ``Evolutionary Question Generation.''}
\label{fig:case2}
\end{figure*}

\subsection{Case Study}
\label{sec:case_study}

The case studies illustrate how TAP-RAG uses one policy interface for different evidence behaviors. Figure~\ref{fig:case1} shows a table-cell extraction case from case1.pdf. The query asks for Novo Nordisk's wages and salaries in 2020. The table contains adjacent values for pension costs, social security contributions, and total staff costs, so semantic retrieval can select a related but wrong number. TAP-RAG treats the query as value extraction, uses TA-QFD to retrieve table-centered evidence, and uses TAVE to verify the row-column intersection. This prevents nearby numeric distractors from being accepted merely because they are semantically close to the query. The final answer, DKK 26,778 million, is accepted after structural and visual evidence agree.

\vspace{\baselineskip}

Figure~\ref{fig:case2} shows a section-level aggregation case from case2.pdf. The query asks which construction step has the longest textual description, with the answer Section 2.3, ``Evolutionary Question Generation.'' The challenge is comparing neighboring sections under a bounded scope. TAP-RAG routes the query as aggregation-oriented, expands within the relevant section range, and selects the answer after support-aware comparison. Thus, the system does not stop at the first high-overlap section, but compares candidate sections according to the requested property.

Together, the cases show why TAP-RAG treats long-document multimodal QA as controlled evidence use. The first needs local structural precision and visual validation; the second needs broader but bounded aggregation. TAPC links both through policy fields for locality, visual acquisition, fusion authority, structural operators, and support thresholds, allowing TA-QFD and TAVE to work as coordinated executors rather than independent retrieval add-ons. The examples also expose distinct baseline failure modes, making the policy decisions and evidence paths easier to audit and compare.

\vspace{-0.5\baselineskip}
\section{Conclusion}
\label{sec:conclusion}
\vspace{0.1\baselineskip}

We presented TAP-RAG, a task-aware policy-controlled framework for long-document multimodal QA. TAPC maps each query into an executable policy controlling graph diffusion, visual acquisition, fusion, structural operators, support checking, and abstention. This makes RAG behavior explicit and auditable: questions over the same document can use different retrieval scopes, inspection strategies, support thresholds, and refusal rules.

Experiments on DocBench and MMLongBench-Doc show gains across domains and page ranges, including long-document settings. The same-backbone comparison with RAG-Anything shows that improvements come from query-time evidence control, not a stronger generator. Ablations show that TA-QFD and TAVE are most effective with TAPC and guarded synthesis.

These results suggest treating long-document multimodal QA as controlled evidence use rather than retrieval scaling. TAP-RAG offers an interpretable policy interface, while future work can improve policy calibration, visual region selection, and task diagnostics for reliable document-level reasoning. These observations emphasize that TAP-RAG improves reliability not by adding context, but by deciding when, where, and how evidence should contribute safely. This direction supports stronger human oversight when document evidence is incomplete or internally conflicting, especially in high-stakes document reasoning scenarios.

\clearpage

\section*{Limitations}

TAP-RAG is evaluated on DocBench and MMLongBench-Doc with a fixed parser, judge protocol, and LLM/VLM backend, so results may vary with different document parsers, models, or benchmark distributions. The task category set covers common long-document multimodal QA behaviors, but unusual layouts, languages, or domain conventions may require new categories or recalibration.

This paper reports domain-level, length-level, ablation, and qualitative analyses. Because the core claim concerns task-specific evidence behavior, task-type-level accuracy would be a valuable additional diagnostic when reliable task labels are available. The current results should therefore be interpreted together with ablations and case studies rather than from the domain table alone.

Finally, the policy fields are manually designed rather than learned end-to-end. This improves interpretability and auditability, but may miss dataset-specific strategies that a learned controller could discover.

\section*{Ethical Considerations}

This work studies document question answering on benchmark-style documents and does not introduce new human-subject data. The main ethical concern is unsupported answer generation in document analysis workflows. TAP-RAG includes evidence support validation and abstention, but these mechanisms are not guarantees in high-stakes settings such as legal, financial, or medical decision making. Practical deployments should preserve document privacy, respect data-use restrictions, avoid storing secrets in environment files committed to repositories, and require appropriate human review for consequential decisions.

\noindent\textbf{Use of AI Assistants.}
AI assistants were used only for language polishing, LaTeX formatting, and checklist-writing assistance. All scientific content, experiments, results, and final manuscript decisions were reviewed and approved by the authors.
\bibliography{custom}

\appendix

\section{Executable Runtime Configuration}
\label{app:runtime_config}

This appendix reports the executable settings used in the reported TAP-RAG run. The configuration is derived from the runtime scripts and query-time policy modules. Environment variables are resolved at startup, legacy aliases are normalized, and required model and embedding credentials are checked before evaluation. The run uses a fixed language/vision backbone, deterministic main generation, TAPC-based policy control, selective visual acquisition, and bounded evidence budgets.

\begin{table*}[!t]
\centering
\begingroup
\compacttablefont
\setlength{\tabcolsep}{3.2pt}
\renewcommand{\arraystretch}{1.05}
\begin{tabularx}{\textwidth}{@{}L{2.8cm}L{3.2cm}Y@{}}
\toprule
Group & Setting & Value used in the reported run \\
\midrule
Backbone & LLM/VLM & Qwen3-VL-Plus for both language generation and page-image vision calls. \\
Backbone & Embedding & text-embedding-v3 with 1024 dimensions. The embedding dimension is fixed to keep indexing and retrieval consistent. \\
Generation & Main temperature & 0 for the main answer path; sampling is used only inside TAPC voting and routing. \\
Parsing & Parser & MinerU with automatic parsing. Image, table, and equation processing are enabled to construct multimodal document chunks. \\
TAPC voting & Samples / temperature / agreement & $K=5$, sampling temperature 0.7, and agreement threshold 0.6. \\
TAPC voting & Validation switches & Structural validation, field coherence, query-document consistency, and final task-decision checking are enabled. \\
TAPC policy & Task policy table update & The Task Policy Table provides task-level defaults, and bounded query control applies visual, local, and global evidence signals. \\
TA-QFD & Seeds and output & Seed top-$k=6$, output top-$k=10$, and include seed nodes disabled. The evidence expansion budget is controlled by the resolved task policy. \\
Retrieval & Mode and budgets & Hybrid mode; retrieval top-$k=8$, semantic top-$k=8$, structural top-$k=12$, and candidate limit 20. \\
Local context & Context budget & Top-8 local contexts, maximum 6000 characters, and maximum 12 files. \\
TAVE & Visual budget & Maximum 2 pages, page window 1, and page-image cache enabled. Visual calls are triggered selectively by the policy. \\
Synthesis & Response behavior & Guarded synthesis produces a concise answer when support is sufficient and abstains when the document does not support the requested fact. \\
Batching & Runtime & Batch size 1, QA concurrency 3, and resume-completed-documents enabled. \\
\bottomrule
\end{tabularx}
\endgroup
\caption{Executable runtime settings used in the reported TAP-RAG evaluation.}
\label{tab:runtime_config_app}
\end{table*}

\section{Task-Aware Runtime Flow}
\label{app:runtime_flow}

TAP-RAG processes each query with a task-aware runtime flow rather than applying one fixed retrieval-and-generation behavior to all questions. The pipeline first resolves document scope, normalizes runtime flags, and constructs a task state containing the predicted task type, target, relation, expected answer form, page hints, visual requirement, locality signal, globality signal, and structural-operator permissions.

The first stage is TAPC. Inside TAPC, multiple sampled LLM routers assign the query to one of seven TPT categories: metadata lookup, evidence localization, document aggregation, value extraction, visual reasoning, Boolean verification, or generic reasoning. Each sample also emits auxiliary evidence signals that estimate whether the query requires visual inspection, local page evidence, or broader document traversal. The final task prior is selected by weighted vote concentration, while the evidence signals are averaged to form the query-time policy inputs.

The Task Policy Table provides a stable default behavior for each category. Metadata lookup emphasizes title-page and header evidence; evidence localization prefers local page-specific evidence; document aggregation permits broader traversal; value extraction enables table or numeric operators; visual reasoning allows selective page-image inspection; and Boolean verification uses stricter support requirements. These defaults make the controller interpretable and prevent runtime behavior from depending only on free-form generation.

Bounded query control then applies constrained adjustments to the table policy. A high visual signal lowers the threshold for TAVE page inspection. A high local signal keeps retrieval close to the most relevant page or section. A high global signal allows broader TA-QFD evidence expansion. These adjustments can tune thresholds, budgets, and fusion behavior, but they do not override hard safety rules such as document scope, support checking, or abstention.

The evidence acquisition stage combines text retrieval, structural evidence, optional TA-QFD expansion, and optional TAVE page inspection. The final response stage combines text, structural, and visual evidence under the selected policy. Visual evidence can validate or supplement the text path, but it is not allowed to freely overwrite a well-supported textual answer.

\section{Framework Prompt Templates}
\label{app:framework_prompts}

Tables~\ref{tab:answer_prompts_app}--\ref{tab:multimodal_prompts_app} summarize the LLM/VLM prompt templates used by TAP-RAG. Placeholders such as query, context, doc\_name, and page\_number are filled at runtime. Implementation-only strings used for deterministic keyword matching, regular-expression checks, file parsing, or environment loading are omitted.

\begin{table*}[!t]
\centering
\begingroup
\compacttablefont
\setlength{\tabcolsep}{3.1pt}
\renewcommand{\arraystretch}{1.05}
\begin{tabularx}{\textwidth}{@{}L{2.7cm}L{3.0cm}Y@{}}
\toprule
Group & Template & Runtime use \\
\midrule
Answer synthesis & Final RAG answer & The assistant answers questions about one specific document. It must be concise, complete, direct, and document-scoped; it starts with the answer, uses only retrieved evidence from the target document, avoids cross-document mixing, and returns ``Not mentioned'' when the requested fact is absent. \\
Answer synthesis & Question-format guidance & The prompt adapts the answer shape to the query type: counts return the exact number plus short item names when useful; page questions return page numbers only; yes/no questions start with ``Yes'' or ``No''; acronym questions return only the expansion; content questions remain short. \\
Document scope & Evidence filter & Given a target document and candidate evidence, the prompt separates valid current-document evidence from cross-document contamination. Clearly unrelated candidates are removed, while uncertain candidates are retained to avoid discarding useful evidence prematurely. \\
Support checking & Evidence sufficiency check & The prompt checks whether the available evidence supports the requested answer. It favors supported concise answers when evidence is adequate and returns an abstention-style response when the document does not contain the requested fact. \\
Guarded response & Answer consistency guard & The guard rejects changes that alter numbers, page numbers, polarity, author identity, or document scope without stronger evidence. It prevents visually plausible but unsupported replacements. \\
Conciseness & Post-processing rewrite & The prompt shortens an answer without adding information or changing meaning. It removes filler, unnecessary citations, and unnecessary lists, starts directly with the answer, and leaves already concise answers unchanged. \\
Page utilities & Page count and existence & The prompts determine the total number of pages or whether a requested page exists in the current document only. They return only the page count for total-page queries, or a compact JSON answer for page-existence checks. \\
\bottomrule
\end{tabularx}
\endgroup
\caption{Answer synthesis, document-scope control, support checking, and guarded-response prompt templates.}
\label{tab:answer_prompts_app}
\end{table*}

\begin{table*}[!t]
\centering
\begingroup
\compacttablefont
\setlength{\tabcolsep}{3.1pt}
\renewcommand{\arraystretch}{1.05}
\begin{tabularx}{\textwidth}{@{}L{2.7cm}L{3.0cm}Y@{}}
\toprule
Group & Template & Runtime use \\
\midrule
TAPC routing & Voting router & The router classifies each query into exactly one of seven TPT categories: metadata lookup, evidence localization, document aggregation, value extraction, visual reasoning, Boolean verification, or generic reasoning. It returns JSON with the target, relation, exact term, query rewrite, expected answer type, reasoning, page hints, confidence, and visual/local/global evidence signals. \\
TAPC routing & Disagreement arbitration & When TAPC voting candidates disagree, an arbitrator examines the query, vote summary, and candidate reasoning, then returns the final JSON task decision. The selected task must match the query's primary evidence requirement. \\
Metadata lookup & First-page metadata extraction & The prompt extracts title, ordered authors, last author, and corresponding author from first-page context. It returns JSON and avoids guessing when author order or correspondence is not supported. \\
Evidence localization & Page locator & Given the target, relation, and candidate page snippets, the prompt identifies the earliest page that introduces the requested concept. It prefers introduction pages over later mentions and considers only pages from the current document. \\
Document aggregation & Count resolver & Given the exact queried term and current-document text, the prompt returns JSON with the exact occurrence count and a short explanation. It counts only current-document content and excludes clearly contaminated evidence. \\
Value extraction & Numeric and table values & Value-extraction prompts request precise factual values from text or tables. Numeric questions focus on explicitly stated quantities, percentages, scores, or dataset sizes, while table-value questions focus on row-column or cell-level evidence. \\
Boolean verification & Support-oriented judgment & Boolean queries require a direct Yes or No answer when evidence is sufficient. If the required relation is absent or ambiguous, the prompt favors conservative refusal rather than unsupported affirmation. \\
Policy signals & Evidence signals & The routing prompt emits visual, local, and global evidence signals in $[0,1]$. These signals adjust visual-trigger thresholds, fusion mode, evidence-expansion behavior, and abstention strength through deterministic policy updates. \\
\bottomrule
\end{tabularx}
\endgroup
\caption{TAPC routing, structural-operator, and policy-control prompt templates.}
\label{tab:task_prompts_app}
\end{table*}

\begin{table*}[!t]
\centering
\begingroup
\compacttablefont
\setlength{\tabcolsep}{3.1pt}
\renewcommand{\arraystretch}{1.05}
\begin{tabularx}{\textwidth}{@{}L{2.7cm}L{3.0cm}Y@{}}
\toprule
Group & Template & Runtime use \\
\midrule
Multimodal parsing & Image analysis & The image prompt asks an expert visual analyst to produce JSON with a detailed description and entity summary. It describes layout, objects, text, visual relationships, colors, actions, and technical details, optionally using surrounding document context. \\
Multimodal parsing & Table analysis & The table prompt asks for JSON with a detailed table description and entity summary. It analyzes table structure, column headers, key values, trends, statistical patterns, relationships among data elements, and the table's significance in context. \\
Multimodal parsing & Equation analysis & The equation prompt asks for JSON explaining mathematical meaning, variable definitions, operations, functions, application domain, theoretical or physical significance, links to surrounding content, and practical use cases. \\
Multimodal parsing & Generic content analysis & The generic-content prompt asks for JSON describing structure, key information, relationships between components, context, significance, and retrieval-relevant details for non-image, non-table, and non-equation content. \\
Chunk construction & Multimodal chunk text & Image, table, equation, and generic chunk templates serialize modality-specific metadata and enhanced captions into retrieval text, including paths, captions, footnotes, table structure, equations, and generated analysis. \\
Query-time analysis & Modality summaries & Query-side prompts briefly describe image content, summarize table data, explain equations, or analyze generic content so that multimodal evidence can be incorporated into the answer-generation context. \\
TAVE & Visual router and page inspection & The visual router decides whether direct page-image inspection is needed. The page-inspection prompt examines only the provided page image and returns JSON with relevance, support score, answer-oriented snippet, page summary, and evidence strings. \\
Cross-modal fusion & Direct visual answer and fusion & The direct visual-answer prompt answers from a single page image only, returning NOT\_RELEVANT if the page lacks evidence. The fusion prompt merges text-path and visual-path answers, resolves conflicts using more specific evidence, discards cross-document content, and outputs a concise final answer. \\
\bottomrule
\end{tabularx}
\endgroup
\caption{Multimodal parsing, TAVE page inspection, and cross-modal fusion prompt templates.}
\label{tab:multimodal_prompts_app}
\end{table*}

\section{Page-Range Values}
\label{app:page_values}

Table~\ref{tab:page_values_app} reports the numerical values used for the page-range accuracy visualization in Figure~\ref{fig:page_range}. Bold values indicate the best result in each page-range column.

\begin{table*}[!t]
\centering
\begingroup
\compacttablefont
\setlength{\tabcolsep}{4.0pt}
\renewcommand{\arraystretch}{1.03}
\begin{tabularx}{\textwidth}{@{}L{3.2cm}*{5}{Z}@{}}
\toprule
\multicolumn{6}{c}{DocBench} \\
\midrule
Method & 1--10 & 11--50 & 51--100 & 101--200 & 200+ \\
\midrule
MMGraphRAG & 63.1 & 62.0 & 58.2 & 54.6 & 55.0 \\
RAG-Anything & 65.1 & \textbf{62.4} & \textbf{61.3} & 68.2 & 68.8 \\
TAP-RAG & \textbf{69.32} & 60.93 & 59.21 & \textbf{68.97} & \textbf{72.73} \\
\midrule
\multicolumn{6}{c}{MMLongBench-Doc} \\
\midrule
Method & 1--10 & 11--50 & 51--100 & 101--200 & 200+ \\
\midrule
MMGraphRAG & 13.8 & 40.8 & 31.3 & 34.2 & 40.0 \\
RAG-Anything & 16.5 & 44.2 & 40.6 & 42.1 & 60.0 \\
TAP-RAG & \textbf{22.4} & \textbf{48.6} & \textbf{45.9} & \textbf{51.2} & \textbf{65.3} \\
\bottomrule
\end{tabularx}
\endgroup
\caption{Page-range accuracy values used in Figure~\ref{fig:page_range}. Bold values mark the best result in each column.}
\label{tab:page_values_app}
\end{table*}

\begin{figure*}[!t]
\centering
\begin{minipage}{0.48\textwidth}
\centering
\begin{tikzpicture}
\begin{axis}[
  ybar,
  width=\linewidth,
  height=0.58\linewidth,
  ymin=0,
  ymax=400,
  ytick={0,100,200,300,400},
  ylabel={QA Pair Count},
  xlabel={Page Range},
  symbolic x coords={1--10,11--50,51--100,101--200,200+},
  xtick=data,
  bar width=7pt,
  enlarge x limits=0.13,
  tick label style={font=\scriptsize},
  label style={font=\small},
  xticklabel style={font=\scriptsize, rotate=0, anchor=center},
  grid=major,
  grid style={gray!25},
  clip=true,
  nodes near coords,
  every node near coord/.append style={font=\scriptsize, yshift=2pt},
]
\addplot[draw=none, fill=orange!65] coordinates {
  (1--10,330)
  (11--50,340)
  (51--100,165)
  (101--200,150)
  (200+,80)
};
\end{axis}
\end{tikzpicture}
(a) DocBench
\end{minipage}\hfill
\begin{minipage}{0.48\textwidth}
\centering
\begin{tikzpicture}
\begin{axis}[
  ybar,
  width=\linewidth,
  height=0.58\linewidth,
  ymin=0,
  ymax=850,
  ytick={0,200,400,600,800},
  ylabel={QA Pair Count},
  xlabel={Page Range},
  symbolic x coords={1--10,11--50,51--100,101--200,200+},
  xtick=data,
  bar width=7pt,
  enlarge x limits=0.13,
  tick label style={font=\scriptsize},
  label style={font=\small},
  xticklabel style={font=\scriptsize, rotate=0, anchor=center},
  grid=major,
  grid style={gray!25},
  clip=true,
  nodes near coords,
  every node near coord/.append style={font=\scriptsize, yshift=2pt},
]
\addplot[draw=none, fill=purple!55] coordinates {
  (1--10,15)
  (11--50,740)
  (51--100,150)
  (101--200,150)
  (200+,110)
};
\end{axis}
\end{tikzpicture}
(b) MMLongBench-Doc
\end{minipage}
\caption{QA-pair distribution by document page range for DocBench and MMLongBench-Doc. The x-axis labels are normalized to the same page-range buckets used in Table~\ref{tab:page_values_app}.}
\label{fig:qa_distribution_app}
\end{figure*}
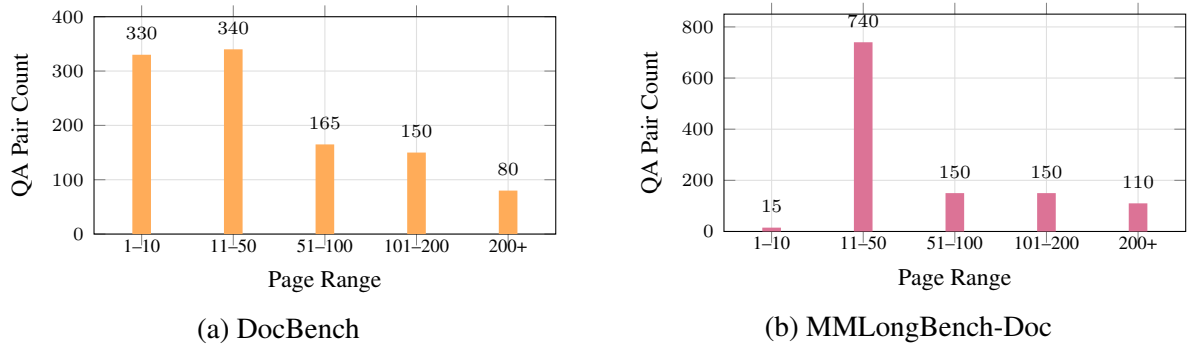

Figure~\ref{fig:qa_distribution_app} reports the QA-pair distribution across page-range buckets for the two benchmarks. These counts contextualize the page-range accuracy values in Table~\ref{tab:page_values_app} and Figure~\ref{fig:page_range}: DocBench is relatively more balanced across short and medium documents, while MMLongBench-Doc concentrates most QA pairs in medium-length documents.

\end{document}